\documentclass[10pt,twocolumn,letterpaper]{article}

\usepackage[pagenumbers]{cvpr} 

\usepackage[dvipsnames]{xcolor}

\usepackage[pagebackref,breaklinks,colorlinks,citecolor=green]{hyperref}

\usepackage{multirow}
\usepackage{comment}
\usepackage{array}
\usepackage{amsfonts,amssymb}
\usepackage{colortbl}
\usepackage{amsmath}

\usepackage{appendix}

\title{SyCoCa: Symmetrizing Contrastive Captioners with Attentive Masking for Multimodal Alignment}

\author{
Ziping Ma\thanks{Work done during an internship with Ant Group.}, \quad Furong Xu, \quad Jian Liu, \quad Ming Yang, \quad Qingpei Guo\thanks{Corresponding author.} \\
Ant Group\\
{\tt\small maziping.im@gmail.com, booyoungxu.xfr@antgroup.com, rex.lj@antgroup.com} \\ 
{\tt\small m.yang@antgroup.com, qingpei.gqp@antgroup.com}
}

\begin{document}
\maketitle

\begin{abstract}
	Multimodal alignment between language and vision is the fundamental topic in current vision-language model research. Contrastive Captioners (CoCa), as a representative method, integrates Contrastive Language-Image Pretraining (CLIP) and Image Caption (IC) into a unified framework, resulting in impressive results. CLIP imposes a bidirectional constraints on global representation of entire images and sentences. Although IC conducts an unidirectional image-to-text generation on local representation, it lacks any constraint on local text-to-image reconstruction, which limits the ability to understand images at a fine-grained level when aligned with texts. To achieve multimodal alignment from both global and local perspectives, this paper proposes Symmetrizing Contrastive Captioners (SyCoCa), which introduces bidirectional  interactions on images and texts across the global and local representation levels. Specifically, we expand a Text-Guided Masked Image Modeling (TG-MIM) head based on ITC and IC heads. The improved SyCoCa can further leverage textual cues to reconstruct contextual images and visual cues to predict textual contents. When implementing bidirectional  local interactions, the local contents of images tend to be cluttered or unrelated to their textual descriptions. Thus, we employ an attentive masking strategy to select effective image patches for interaction. Extensive experiments on five vision-language tasks, including image-text retrieval, image-captioning, visual question answering, and zero-shot/finetuned image classification, validate the effectiveness of our proposed method.
\end{abstract}

\section{Introduction}
\label{sec:intro}
\begin{figure*}[t]
	\centering
	\includegraphics[width=0.9\linewidth]{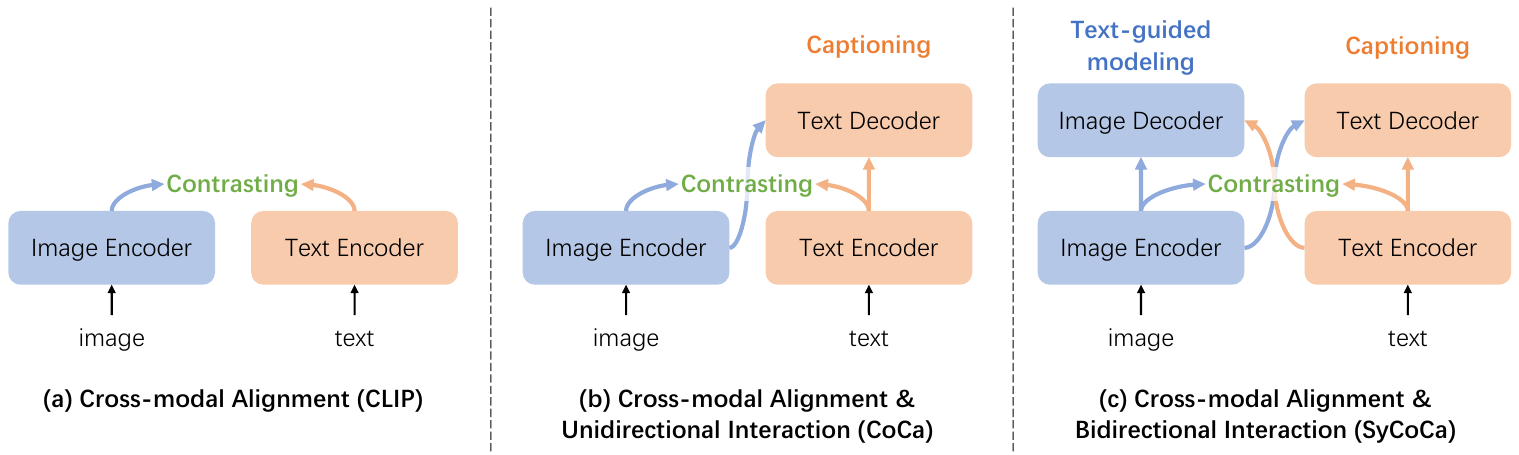}
	
	\caption{Comparison of the pipelines in vision-language pretraining frameworks. (a) CLIP: aligning global features across modalities through contrastive learning. (b) CoCa: introducing image captioning to construct unidirectional fine-grained interaction. (c) Our SyCoCa: bidirectional local interation with attentive masking to enhance comprehensive cross-modal understanding.}
	\label{fig:overview}
\end{figure*}

In recent years, the dramatic progress of multimodal alignment between vision and language has reshaped computer vision research in some sense. The availability of large-scale datasets and powerful computational resources has led to many seminal works and impressive results in this field. Most methods use contrastive objective to constrain global representation between modalities as shown in Figure~\ref{fig:overview} (a), such as CLIP \cite{radford2021learning,yang2022chinese}, ALBEF \cite{li2021align}, mPLUG \cite{li2022mplug,xu2023mplug}. BEiT-v3 \cite{wang2023image} treats image as a foreign language and uses a  mask-then-predict strategy for pre-training. This pretraining method requires separate fine-tuning for each downstream task and the pretrained model cannot be directly used across tasks. CoCa \cite{coca} combines Image-Text Contrastive (ITC) and Image Caption (IC), resulting in a pretrained model that can be directly used for both retrieval and IC tasks. It is a promising architecture and works well on multiple vision-language tasks.

However, in terms of local interaction between image patches and text tokens, the IC head in CoCa only utilizes the visual cues to generate textual descriptions, yet disregarding the visual context reconstruction from textual cues.  From vision pre-training works, such as SimMIM \cite{xie2022simmim} and  MAE \cite{he2022masked}, we learn that the image reconstruction can learn a strong content representation. Therefore, in the multimodal scenario, when text cue is introduced into the image reconstruction task and local interaction is performed, the representation of texts and images can be unified into one space, thereby further enhancing multimodal alignment.

In this paper, we propose a novel framework called Symmetrizing Contrastive Captioners (SyCoCa) that incorporates both local image-to-text generation and text-to-image reconstruction in addition to global constractive objective. In addition to ITC head and IC head, we introduce a text-guided masked image modeling (TG-MIM) head. The difference between CoCa and our method is illustrated in Figure~\ref{fig:overview}. CoCa only achieved undirectional interaction between image and text as shown in Figure~\ref{fig:overview} (b). Further, in Figure~\ref{fig:overview} (c), our TG-MIM introduces text for image reconstruction, which enhances the fine-grained representation ability of images. In this way, SyCoCa has the bidirectional global and local interactions between modalities.

During the actual multimodal alignment process, although an image is worth one thousand words, people seldom write one thousand words to describe the image content. Instead, text descriptions or image captions are often highly abstract just focusing on the main character, object, or event in an image. For example, the caption of a picture about the family dinner of Thanksgiving would rarely delineate the furniture in the room in details. Moreover, even short text descriptions may imply rich and abundant image contents. \textit{e.g.,}. \emph{a football match} may well imply a vivid scene of a grass field, multiple palyers, and croweded audience, etc. Therefore, it is necessary to choose appropriate local regions and representation for alignment. To select effective image patches for text, we employ an attentive masking strategy. Specifically, we calculate the similarity between image tokens and text tokens to determine the relevance of image patches. For the IC task, the most pertinent image patches are selected to aid in generating textual descriptions by considering their semantic similarity with text descriptions. In contrast, we choose the least relevant image patches with the text on the TG-MIM task, aiming to leverage the text tokens to assist in recovering image content. Extensive experiments demonstrate the effectiveness of our proposed method on five downstream tasks. 

In summary, the main contributions of our work are listed as follows:

\begin{itemize}
	\item We first propose a symmetrizing contrastive captioners for multimodal alignment, which improves the understanding between images and sentences from a global and local perspective. 
	\item To promote bidirectional local interaction, we adopt an attentive masking strategy to select appropriate image patches for IC and TG-MIM heads respectively.
	\item Thorough experiments validate that our proposed \textit{SyCoCa} outperforms CoCa on several downstream tasks, {\it e.g.} image-text retrieval,  image-captioning, visual question answering and  image classification. For example, we obtained +5.1$\%$/3.7$\%$ in mTR/mIR on Flicker-30k compared to CoCa in image-text retrieval tasks.
\end{itemize}

\section{Related Work}
\label{sec:related}

\textbf{Vision-Language Pretraining.}
In recent years, there has been tremendous progress in multimodal alignment, especially between vision and language. Extensive researchers have dedicated their efforts to exploring vision-language pretraining. Early works \cite{tan2019lxmert,chen2020uniter,zhang2021vinvl} prefixed a pretrained object detection modules to extract visual representations, which were then aligned with textual representations to achieve multimodal alignment. Later efforts focused on training multimodal transformers from scratch, such as ViLT \cite{kim2021vilt} and VLMo \cite{bao2022vlmo}.  Pre-training of a foundation model on gigantic data with tremendous computation led to the breakthrough of multimodal alignment of images and texts. CLIP \cite{radford2021learning} and ALIGN \cite{jia2021scaling} trained a dual-encoder on large-scale noisy image-text pairs using contrastive loss, obtaining generic image and text representations for crossmodal alignment and zero-shot image classification tasks. Florence \cite{yuan2021florence} used a unified contrastive learning for various vision and image-text benchmarks. BLIP \cite{li2023blip} trained a Q-former to align a frozen vision encoder and language encoder. BEiT-v3 \cite{wang2023image} treated image as a foreign language, and used mask-then-predict strategy for pre-training. To enhance local interaction of images and texts, CoCa \cite{coca} introduced a decoder for image caption based on CLIP. Our SyCoCa further achieves bidirectional local interactions and enhances fine-grained alignment capabilities between modalities.

\textbf{Masked Image Modeling.} 
Beyond contrastive learning, masked image modeling (MIM) \cite{chen2020generative,doersch2015unsupervised,doersch2015unsupervised,pathak2016context,trinh2019selfie} emerges as another promising technique in vision pre-training. Recently, iGPT \cite{chen2020generative}, ViT \cite{dosovitskiy2020image} and BEiT \cite{bao2021beit} recalled this learning approach on the modern vision transformers, which show great potential in representation learning by introducing special designs, such as clustering on pixels \cite{chen2020generative}, prediction of mean color \cite{dosovitskiy2020image}, and tokenization via an additional dVAE network with a block-wise masking strategy \cite{bao2021beit}. SimMIM \cite{xie2022simmim} predicted RGB values of raw pixels by direct regression. MAE \cite{he2022masked} used an asymmetric
encoder-decoder architecture, where the encoder operates only on the visible subset of patches, then a lightweight decoder reconstructs
the original image from the latent representation and mask tokens. 

Inspired by MAE, we propose a TG-MIM head to improve fine-grained multimodal alignment. The major differences of TG-MIM against MAE include: 1) the text cues are leveraged in TG-MIM to guide image reconstruction; 2) an attentive masking scheme is used in TG-MIM to select appropriate patches; and 3) TG-MIM targets on improving multi-modal alignment, instead of learning vision representation as in MAE. 

\section{Method}
\label{sec:method}

We are interested in leveraging bidirectional cross-modal interaction to learn fine-grained visual and textual representations within an aligned latent space. Contrastive Captioner (CoCa) proposes to incorporate an image captioning task alongside the contrastive task, which learns fine-grained representation and establishes a stronger alignment between modalities than CLIP. It is worth noting that the modality interaction in CoCa is unidirectional and asymmetric, focusing solely on genetating text from images.

\begin{figure*}[t]
	\centering
	\includegraphics[width=0.75\linewidth]{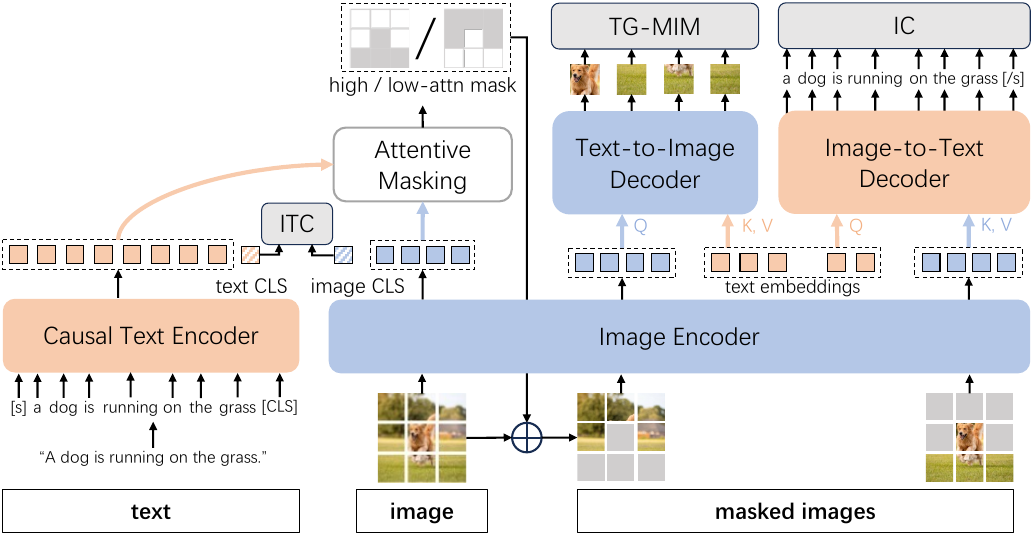}
	
	\caption{The detailed illustration of our proposed method. The framework of our method consist of four modules: an image encoder, a (causal) text encoder, a (text-to-)image decoder, a (image-to-)text decoder. Our method focuses on three pretraining objectives: image-text contrasting (ITC), text-guided masked image modeling (TG-MIM) and image captioning (IC).}
	\label{fig:model}
\end{figure*}

To enhance vision-language alignment, we introduce a novel framework called SyCoCa, which expands the existing CoCa model by incorporating bidirectional local interaction with attentive masking. Our framework comprises of three objectives: (i) image-text contrasting (ITC), (ii) image captioning (IC), and (iii) text-guided masked image modeling (TG-MIM). Figure~\ref{fig:model} shows the overall framework of SyCoCa.

The goal of SyCoCa is to further explore the potential of cross-modal interactive prediction tasks. Building upon CoCa, SyCoCa incorporates TG-MIM to establish predictions from text to image, aiming to compensate for the inherent asymmetry in CoCa. This establishes bidirectional prediction between modalities, promoting fine-grained vision-language understanding. Furthermore, we design a novel attentive masking procedure that enables the bidirectional cross-modal interactions to focus on different regions of images, guided by the accompanying texts.

\subsection{Model Architecture}

As shown in Figure~\ref{fig:model}, the overall model architecture consists of four key components: the image encoder, text encoder, (text-to-)image decoder, and (image-to-)text decoder. Here, we provide a detailed explanation to each ones. 

\textbf{Image Encoder.} We employ vision transformer as the image encoder to model an input image. The image encoder takes image patches as input and encodes them into a sequence of embeddings $\{ v^{cls}, v^1, ..., v^P\}$ where each embedding corresponds to a specific image patch. Additionally, an extra $\mathtt{[CLS]}$ embedding is included to provide a global representation of the image.

\textbf{Text Encoder.} As the text encoder, we adopt a causal masked transformer encoder to model text inputs. This encoder takes the simple BPE tokenized input text and converts it into a sequence vector represented as $\{w^1, ..., w^S, w^{cls}\}$, in which the embedding of the $\mathtt{[CLS]}$ token summarizes the global text feature. The purpose of adopting causal attention mask is to prevent any potential information leakage from future tokens to past tokens during the text encoding process.

\textbf{Image / Text Decoder.} To further capture the interaction between image-text pairs, we use an image decoder and a text decoder. Each decoder utilizes cross-attention transformer modules to deeply fuse image and text information, enabling bidirectional local interaction. The cross-modality multi-head attention module uses the representations of one modality as the query and the other modality’s representations as the key and value. More specifically, the image decoder predicts the pixel values of masked image patches, leveraging cross-attention to incorporate relevant text. Similarly, the text decoder predicts the logits of the next token, utilizing cross-attention to integrate relevant image information. This bidirectional fusion mechanism ensures a comprehensive mutual understanding of the image-text pairs.

\subsection{Training objectives}

We start from an image-text pair dataset consisting of pairs $(I_i, T_i)$, where $I_i$ represents an image and $T_i$ is the corresponding text caption. The image encoder $E_I$ and text encoder $E_T$ are responsible for encoding the image and text, respectively, yielding embeddings:
\begin{align}
	& \{v_i^{cls}, v_i^1, ..., v_i^P\} = E_I (I_i) \quad \text{and} \\
	& \{w_i^1, ..., w_i^S, w_i^{cls}\} = E_T (T_i).
\end{align}
$P$ refers to the number of patches, and $S$ is the length of tokenized text sequence. Here, we present the applied training objectives in details.

\textbf{Image Text Contrasting (ITC).} In the training process, we consider a batch of $N$ image-text pairs $\{I_i, T_i\}^N_{i=1}$ and their corresponding global representations $\{v_i^{cls}, w_i^{cls}\}^N_{i=1}$. It is assumed that these representations have been normalized to have a unit $\ell_2$ norm. The contrasive objective is designed to align the image and text representations:
\begin{align}
	\mathcal{L}_{ITC} &= \frac{1}{2N} \left[ \sum^N_{i} \log \left( \frac{\exp(\langle v^{cls}_i, w^{cls}_i\rangle/\tau)}{\sum^N_{j=1}\exp(\langle v^{cls}_i, w^{cls}_j \rangle/\tau)} \right) \right] \nonumber \\ 
	&+ \frac{1}{2N} \left[ \sum^N_{i} \log \left( \frac{\exp(\langle v^{cls}_i, w^{cls}_i\rangle/\tau)}{\sum^N_{j=1}\exp(\langle v^{cls}_i, w^{cls}_j \rangle/\tau)} \right) \right],
\end{align}
where $\langle \cdot, \cdot \rangle$ refers to the inner product, and $\tau$ is the temperature to scale the logits. The contrastive objective that pulls the representations of paired image-text close together while pushing apart unmatched pairs, promoting alignment in a shared semantic space.

\textbf{Image Captioning.} Image captioning objective requires the model to generate tokenized texts $T_i$ with precise details in an autoregressive manner, compared with ITC treating the inputs as single entities. We use the image encoder $E_I$ to encode image representations and train the text decoder $D_T$ to maximize the conditional likelihood of the text $T_i$ by utilizing forward autoregressive factorization:
\begin{equation}
	\mathcal{L}_{IC} = - \sum_{j=1}^{\left| T_i \right|}\log D_T \left( T_i^j | T_i^{<j}, E_I(I_i) \right).
\end{equation}
The text decoder is trained using parallel computation for enhanced learning efficiency. This training objective enables the model to learn fine-grained representations with strong alignment through cross-modal prediction, fostering the acquisition of joint-modality representations useful for various multimodal understanding and generation tasks. Nonetheless, this interaction is unidirectional and asymmetrical, primarily focusing on understanding visual elements in language, without exploiting the comprehension of textual elements in the visual context.

\textbf{Text-Guided Masked Image Modeling.}
To address the inherent imbalance in cross-modal interaction within the image captioning task, we design a novel training objective called text-guided masked image modeling. Our SyCoCa framework utilizes the image decoder, denoted as $D_I$, to reconstruct the masked image by predicting the pixel values for each masked patch. The output linear projection and reshape process is similar to that of the MAE. For a given image $I = \{ p^1, ..., p^P\}$ consisting of $P$ patches, we employ patch-wise masking using a masking map $M = \{m^1, ..., m^P\}$, where $m^i\in\{0, 1\}$ indicates whether the patch is masked. The TG-MIM loss is computed as the L1 loss between the masked patches and their reconstructed counterparts in the pixel space:
\begin{align}
	& \{\hat{p}^1, ..., \hat{p}^P\} = D_I(E_I(I \oplus M), E_T(T)), \\
	& \mathcal{L}_{TM} = \sum_{i=1}^P m^i \cdot \parallel p^i - \hat{p}^i \parallel ^ 1,
\end{align}
where $\hat{p}^i$ refers to the reconstructed patches. Note that the TG-MIM objective, proposed in this paper, distinguishes itself from the image-pretraining objective MIM by dedicated efforts on the guidance offered by the cross-attended text representation. The text-guided modeling focuses on the interaction between visual and textual representations, with the goal of enhancing the fine-grained understanding of textual elements within the visual context through reconstruction process. The introduction of this objective complements the uni-directional prediction of image captioning.

\subsection{Attentive Masking}

Intuitively, the reconstruction of image regions that reveal the same semantics or show highly related contents to the text shall be a more effective and robust cross-modal interaction, rather than reconstructing those irrelevant image regions. Therefore, we suggest generating masking maps that eliminate image patches that are relevant to language captions, and subsequently reconstruct these patches based on the text captions.

A straightforward idea to evalaute the correlation is by measuring the similarity between each patch embedding $v^i$ and the global text embedding $w^{cls}$. However, $w^{cls}$ provides a coarse-grained summary of the entire sentence. To mask the vision elements mentioned in the caption as accurately as possible, we propose to calculate the token-wise maximum similarity between image and text embeddings. This method is effective in modeling fine-grained cross-modal similarities in a previous study~\cite{yao2021filip}. Formally, for the $i$-th embedding $v^i$ of an image, we calculate its similarities with all text embeddings $\{w^1, ..., w^S\}$ and select the largest similarity
\begin{equation}
	s^i = \max\nolimits_{j=1}^S \langle v^i, w^j \rangle
\end{equation}
to represent its correlation with the text.

\begin{table*}[t]
	\centering
	\resizebox{0.9\linewidth}{!}{
		\begin{tabular}{l|cccccc|cccccc}
			\toprule
			& \multicolumn{6}{c|}{Flickr30K} & \multicolumn{6}{c}{MSCOCO} \\
			\cmidrule(r){2-7} \cmidrule(r){8-13}
			& \multicolumn{3}{c}{Image $\rightarrow$ Text} & \multicolumn{3}{c|}{Text $\rightarrow$ Image} & \multicolumn{3}{c}{Image $\rightarrow$ Text} & \multicolumn{3}{c}{Text $\rightarrow$ Image} \\
			\cmidrule(r){2-4} \cmidrule(r){5-7} \cmidrule(r){8-10} \cmidrule(r){11-13}
			& R@1 & R@5 & R@10 & R@1 & R@5 & R@10 & R@1 & R@5 & R@10 & R@1 & R@5 & R@10 \\
			\midrule
			CoCa & 44.2 & 72.4 & 81.7 & 31.3 & 59.7 & 70.8 & 16.3 & 37.1 & 48.9 & 15.3 & 35.8 & 47.0 \\
			Ours & \textbf{46.6} & \textbf{76.3} & \textbf{83.6} & \textbf{35.9} & \textbf{64.6} & \textbf{75.6} & \textbf{18.7} & \textbf{41.0} & \textbf{53.6} & \textbf{17.2} & \textbf{39.7} & \textbf{51.6} \\
			\midrule
			\%Gains & +5.4 & +5.4 & +2.3 & +14.7 & +8.2 & +6.8 & +14.7 & +10.5 & +9.6 & +12.4 & +10.9 & +9.8 \\
			\bottomrule
		\end{tabular}
	}
	\caption{Zero-shot image-text retrieval evaluation results on Flickr30K and MSCOCO dataset.}
	\label{tab:zero-shot retrieval}
\end{table*}

\begin{table*}[t]
	\centering
	\resizebox{0.75\linewidth}{!}{
		\begin{tabular}{l|cccc|cccc|ccc}
			\toprule
			& \multicolumn{8}{c|}{Image Captioning} & \multicolumn{3}{c}{vision question answering} \\
			\cmidrule(r){2-9} \cmidrule(r){10-12}
			& \multicolumn{4}{c|}{MSCOCO} & \multicolumn{4}{c|}{NoCaps-Test} & \multicolumn{3}{c}{VQA} \\
			\cmidrule(r){2-5} \cmidrule(r){6-9} \cmidrule(r){10-12}
			& B@4 & M & C & S & B@4 & M & C & S & val & test-dev & test \\
			\midrule
			CoCa & 21.3 & 22.6 & 71.1 & 16.2 & 12.7 & 22.4 & 53.4 & 10.4 & 43.0 & 39.1 & 39.6 \\
			Ours & \textbf{22.4} & \textbf{23.6} & \textbf{75.6} & \textbf{16.8} & \textbf{12.9} & \textbf{22.8} & \textbf{54.8} & \textbf{10.5} & \textbf{46.9} & \textbf{42.6} & \textbf{42.9} \\
			\midrule
			\%Gains & +5.2 & +4.4 & +6.3 & +3.7 & +1.6 & +1.8 & +2.6 & +1.0 & +9.1 & +9.0 & +8.3 \\
			\bottomrule
		\end{tabular}
	}
	\caption{Comparison with CoCa on image captioning (MSCOCO, NoCaps) and vision question answering (VQA). B: BLEU@4. M: METEOR. C: CIDEr. S: SPICE.}
	\label{tab:caption and vqa}
\end{table*}

\begin{figure*}[t]
	\centering
	\includegraphics[width=0.88\linewidth]{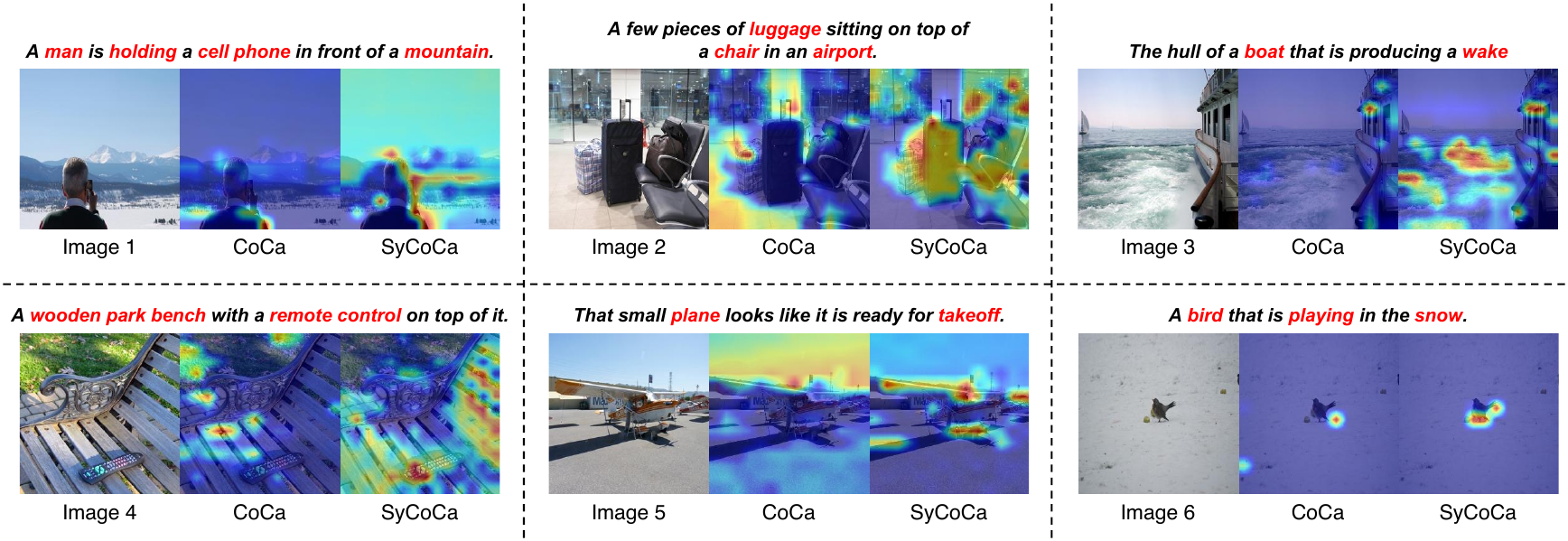}
	\caption{Qualitative analysis of the proposed SyCoCa. We visualize the attention localization map of the first convolution layer in image encoder by the toolkit Grad-CAM.}
	\label{fig:visualization}
\end{figure*}

\begin{table*}[t]
	\centering
	\resizebox{0.9\linewidth}{!}{
		\begin{tabular}{l|cccccccc|c}
			\toprule
			& IN-1K~\cite{imagenet} & IN-V2~\cite{imagenet-v2} & IN-A~\cite{imagenet-a} & IN-R~\cite{imagenet-r} & C-10~\cite{cifar} & C-100~\cite{cifar} & STL-10~\cite{stl-10} & Caltech~\cite{caltech101} & Avg. \\
			\midrule
			CoCa & 26.1 & 22.4 & 6.5 & 31.7 & \textbf{65.9} & 29.7 & 88.1 & 64.5 & 41.9 \\
			Ours & \textbf{27.8} & \textbf{23.9} & \textbf{7.0} & \textbf{32.3} & 62.0 & \textbf{29.8} & \textbf{89.2} & \textbf{65.6} & \textbf{42.2} \\
			\midrule
			\%Gains & +6.5 & +6.7 & +7.7 & +1.9 & -5.9 & +0.3 & +1.2 & +1.7 & +2.5 \\
			\bottomrule
		\end{tabular}
	}
	\caption{Zero-shot image classification evaluation results on 8 coarse-grained datasets: ImageNet, ImageNet-V2, ImageNet-A, ImageNet-R, CIFAR-10, CIFAR-100, STL-10 and Caltech101.}
	\label{tab:zero-shot classification}
\end{table*}

\begin{table*}[t!]
	\centering
	\resizebox{0.80\linewidth}{!}{
		\begin{tabular}{l|ccccccc|c}
			\toprule
			& DTD~\cite{dtd} & Dogs~\cite{dogs} & CUB~\cite{cub} & Pets~\cite{pets} & Flowers\cite{flowers} & MNIST~\cite{mnist} & Food~\cite{food101} & Avg. \\
			\midrule
			CoCa & 58.1 & 49.0 & 40.6 & \textbf{72.7} & 69.1 & 93.3 & 69.1 & 64.6 \\
			Ours & \textbf{60.5} & \textbf{51.2} & \textbf{41.3} & 70.0 & \textbf{70.6} & \textbf{94.5} & \textbf{71.1} & \textbf{65.6} \\
			\midrule
			\%Gains & +4.1 & +4.5 & +1.7 & -3.7 & +2.2 & +1.3 & +2.9 & +1.9 \\
			\bottomrule
		\end{tabular}
	}
	\caption{Fine-tuned image classification evaluation results on 7 fine-grained datasets: DTD, Oxford-IIIT Pet, Stanford Dogs, CUB-200, Flowers102, MNIST, Food101. Both models are transfered using linear probing.}
	\label{tab:fine-tuned classification}
\end{table*}

Image patches to be masked are selected based on their scores $\{s^1, ..., s^P\}$. To ensure stable training in TG-MIM, a fixed ratio $r_h$ of top scoring patches (represented by the \textit{high attn mask} in Figure~\ref{fig:model}) in the input image are masked. Conversely, in the caption task, the patches with the lowest scores (represented by the \textit{low attn mask} in Figure~\ref{fig:model}) in the input image are masked at a fixed ratio $r_l$. The purpose is to enhance cross-modal interaction by encouraging the model to prioritize visually salient regions that play a significant role in understanding and representing the image for caption generation.

\subsection{Overall objective}
Finally, we pretrain SyCoCa with all these losses combined:
\begin{equation}
	\mathcal{L} = \mathcal{L}_{ITC} + \lambda_{IC}\mathcal{L}_{IC} + \lambda_{TM}\mathcal{L}_{TM},
\end{equation}
where $\lambda_{IC}$ and $\lambda_{TM}$ are the hyper-parameters weighting between IC and TG-MIM. All the modules of SyCoCa are trained from scratch.

\section{Experiments}
\label{sec: experiments}

\begin{table*}
	\centering
	\resizebox{0.9\textwidth}{!}{
		\begin{tabular}{l|cccccc|cccc|cccc|c}
			\toprule
			& & & & & & & \multicolumn{4}{c|}{zero-shot retrieval} & \multicolumn{4}{c|}{zero-shot classification} & \\
			\cmidrule(r){8-11}\cmidrule(r){12-15}
			& & & & & & & \multicolumn{2}{c}{Flickr30K} & \multicolumn{2}{c|}{MSCOCO} & \multirow{2}{*}{IN-1K} & \multirow{2}{*}{C-10} & \multirow{2}{*}{STL-10} & \multirow{2}{*}{Caltech} & \\
			\cmidrule(r){8-9}\cmidrule(r){10-11}
			& ITC & IC & MIM & TG-MIM & RM & AM & mTR & mIR & mTR & mIR & & & & & Avg.\\
			\midrule
			& $\checkmark$ &  &  &  &  &  & 32.0 & 24.4 & 13.5 & 14.8 & 10.7 & 41.0 & 68.4 & 36.7 & 30.2 \\
			\rowcolor{blue!10} CoCa & $\checkmark$ & $\checkmark$ &  &  &  &  & 37.5 & 28.7 & 16.1 & 16.0 & 10.4 & 39.6 & 69.2 & 37.8 & 31.9 \\
			& $\checkmark$ & $\checkmark$ &  &  &  & $\checkmark$ & 41.1 & 31.4 & 17.5 & 18.3 & 11.4 & 40.0 & 71.7 & 39.1 & 33.8 \\
			& $\checkmark$ & $\checkmark$ & $\checkmark$ &  & $\checkmark$ &  & 36.0 & 27.1 & 15.8 & 16.3 & 11.8 & 46.6 & 71.5 & 38.5 & 33.0 \\
			& $\checkmark$ & $\checkmark$ & $\checkmark$ &  &  & $\checkmark$ & 37.7 & 27.8 & 15.2 & 16.2 & 11.6 & 43.5 & 72.1 & 35.9 & 32.5 \\
			& $\checkmark$ & $\checkmark$ &  & $\checkmark$ & $\checkmark$ &  & 35.8 & 26.4 & 14.4 & 15.2 & \textbf{11.9} & 44.8 & 71.4 & 39.2 & 32.4 \\
			\rowcolor{red!10} Ours & $\checkmark$ & $\checkmark$ &  & $\checkmark$ &  & $\checkmark$ & \textbf{42.6} & \textbf{32.4} & \textbf{18.3} & \textbf{18.4} & 11.6 & \textbf{51.1} & \textbf{73.6} & \textbf{39.5} & \textbf{35.9 (+12.5\%)} \\
			\bottomrule
		\end{tabular}
	}
	\caption{Ablation study on training objectives. ITC: image-text contrasting IC: image captioning. MIM: masked image modeling. TG-MIM: text-guided masked image modeling. RM: random masking for MIM and TG-MIM. AM: attentive masking for MIM and TG-MIM. mIR/mTR refers to the corresponding mean value of R@1, R@5 and R@10.}
	\label{tab:object ablation}
\end{table*}

To demonstrate the effectiveness of our proposed SyCoCa, we conduct extensive experiments on 5 downstream tasks. Initially, we present the experimental setup, including model architecture, pretraining datasets, downstream tasks, and implementation details. Subsequently, we compare SyCoCa with the baseline model, CoCa, on image-text retrieval, image classification, image captioning and visual question answering tasks. Finally, we perform a series of ablation studies to further analyze and evaluate our model.

\subsection{Experimental Setup}
\textbf{\quad Model Architecture.} 
To ensure a fair comparison, we conduct our experiment using the open-source implementation of CoCa\footnote{\url{https://github.com/mlfoundations/open_clip}}. Both SyCoCa and the baseline model utilize the same CoCa-Base configuration for the image encoder, text encoder, and text decoder. However, in SyCoCa, we introduce a new image decoder that shares the same architecture as the text decoder yet with minor modifications. Specifically, we replace the token-prediction head with a pixel-prediction head. All model parameters are initialized using a Gaussian distribution with a mean of 0 and a standard deviation of 0.01, allowing the training process to start from scratch.

\textbf{Pretraining Data.} 
We use the Conceptual Captions 12M~\cite{cc12m} (CC12M) dataset with 12 million image-caption pairs, as the multi-modal pretraining data for all models. Although this dataset is smaller in scale compared to the large custom datasets employed by the state-of-the-art pretraining models (such as 400M pairs in CLIP~\cite{radford2021learning} and 3B pairs in CoCa~\cite{coca}), it is a good match to the computation resources available to us. Moreover, CC12M has been widely adopted for benchmark evaluations in various studies on vision-language pretraining~\cite{slip, flava, cyclip, glip, lit}.

\textbf{Downstream Tasks.} We evaluate SyCoCa on 3 downstream vision-language tasks and 2 classification tasks, including image-text retrieval, image captioning, visual question answering, zero-shot image classification, and fine-tuned image classification.

\textbf{Implementation Details.} We conduct our model training on two machines, each equipped with 8 NVIDIA A100 GPUs, for a total of 20 epochs. The batch size during training is set to 2048, and the resolution of pretraining images is set to 224$\times$224. We use the AdamW optimizer~\cite{adamw} with an initial learning rate of $1e-4$. The learning rate schedule follows a cosine decay, including a warm-up period of 5000 steps. In terms of hyperparameters, we simply set $\lambda_{IC} = 2$ following CoCa and $\lambda_{TM} = 1$. The masking ratios $r_h$ and $r_l$ are both empirically set to $50\%$.

\subsection{Results on Downstream Vision-Language Tasks}

We compare the performance of SyCoCa and CoCa on downstream vision-language tasks, including image-text retrieval, image captioning, and vision question answering.

\textbf{Image-Text Retrieval.} In this task, the models are required to find the sample that best matches the input across modalities without finetuning. We conduct evaluation on standard image-text retrieval datasets, namely Flickr30K~\cite{flickr30k} and MSCOCO~\cite{mscoco}, and report the results in~\ref{tab:zero-shot retrieval}. We can find that SyCoCa consistently outperforms CoCa showcasing the gains in the range of $5\%$-$15\%$ for R@1.

\textbf{Image Captioning.} In this task, the models are required to generate textual descriptions for input images. We fine-tune both SyCoCa and CoCa using cross-entropy loss on the MSCOCO Captioning~\cite{mscoco} dataset. Subsequently, we report the BLEU@4, METEOR, CIDEr, and SPICE scores on the Karparthy test split of MSCOCO, as well as the test split of the NoCaps~\cite{nocaps} dataset. The results shown in Table~\ref{tab:caption and vqa} demonstrate that SyCoCa outperforms CoCa across all metrics. Specifically, SyCoCa improves by $4\%$-$6\%$ on MSCOCO and $1\%$-$3\%$ on NoCaps compared to CoCa.

\textbf{Vision Question Answering.} In this task, the models are required to predict an answer based on both an image and a question. To apply SyCoCa and CoCa to this task, we modify the text encoder to a multi-modal decoder, which allows for the fusion of image representation and question input. We adapt the text decoder to generate answers in an autoregressive manner, derived from the output of the multi-modal decoder. To finetune the models, we utilize the VQA~\cite{vqa} dataset. During inference, we constrain the decoder to generate answers only from a set of 3,128 candidate answers to make a fair comparison. The results depicted in Table~\ref{tab:caption and vqa} clearly indicate that SyCoCa surpasses CoCa in all cases, where SyCoCa achieves remarkable improvements of $8\%$-$9\%$ on the validation, test-dev, and test splits of VQA.

Our evaluation on downstream vision-language tasks has confirmed the advancements achieved by SyCoCa in terms of multi-modal alignment and cross-modal understanding. On the one hand, our zero-shot retrieval experiments have demonstrated the importance of bidirectional mutual interaction in enhancing the performance of CoCa in terms of modality alignment. On the other hand, the experimental results presented in Table~\ref{tab:caption and vqa} highlights the critical role played by the bidirectional interaction mechanism in fostering mutual understanding and capturing fine-grained elements across different modalities.

\subsection{Results on Image Classification Tasks}
We conduct a comparison between CoCa and SyCoCa regarding their classification performance on both zero-shot and fine-tuned image classification tasks. For zero-shot image classification, we evaluate the models on 8 coarse-grained image classification datasets that include common categories like airplanes and dogs. The results are presented in Table~\ref{tab:zero-shot classification}. SyCoCa outperforms CoCa in 7 out of 8 cases, resulting in an average accuracy improvement of 2.5\%. These findings highlight the effectiveness of bidirectional understanding in bridging the gap between visual and textual representations.

In case of the fine-tuned image classification tasks, we use 7 fine-grained image classification datasets that encompass subcategories within a specific categorie, such as different breeds of dogs~\cite{dogs} like Border Collie and Golden Retriever.  During the fine-tuning process, we employ linear probing to gauge the image encoders' capability to discern intricate details within images. We report our results in Table~\ref{tab:fine-tuned classification}. The results indicate that SyCoCa outperforms CoCa in 6 out of 7 datasets, resulting in an average performance improvement of $2\%$. This demonstrates the effectiveness of incorporating bidirectional prediction to enhance the understanding and differentiation of fine-grained elements within the visual representation domain.

\subsection{Qualitive Analysis of SyCoCa}

To obtain a intuitive comprehension of the advantages of SyCoCa, we use Grad-CAM~\cite{selvaraju2017grad}, a commonly used "visual explanation" toolkit, to generate attention location maps for the first convolutional layer in the image encoder. As shown in Figure~\ref{fig:visualization}, compared with CoCa, the improved version SyCoCa can capture fine-grained visual elements related to informative words in text. For instance, in the case of Image 1, SyCoCa exhibits a focused attention towards regions that are pertinent to the words "\textit{holding}", "\textit{cell phone}" and "\textit{mountain}" whereas CoCa neglects these essential details. In contrast, in the case of Image 2, SyCoCa successfully identifies the objects "\textit{luggage}" and "\textit{chair}", as well as the scenario "\textit{airport}", whereas CoCa falls short in recognizing these elements. This enhanced performance is attributed to the fine-grained cross-modal understanding ability enabled by bidirectional local interaction and attentive masking, which proves advantageous for downstream tasks such as image captioning and vision language answering.

\begin{table}
	\centering
	\resizebox{\linewidth}{!}{%
		\begin{tabular}{lll|cccc|cccc|c}
			\toprule
			& & & \multicolumn{4}{c|}{zero-shot retrieval} & \multicolumn{4}{c|}{zero-shot classification} & \\
			\cmidrule(r){4-7} \cmidrule(r){8-11}
			&&& \multicolumn{2}{c}{Flickr30K} & \multicolumn{2}{c|}{MSCOCO} & \multirow{2}{*}{IN-1K} & \multirow{2}{*}{C-10} & \multirow{2}{*}{STL-10} & \multirow{2}{*}{Caltech} & \\
			\cmidrule(r){4-5} \cmidrule(r){6-7}
			& $r_l$ & $r_h$ & mTR & mIR & mTR & mIR &  &  &  & & Avg. \\
			\midrule
			& $25\%$ & $75\%$ & 40.1 & 31.0 & 17.5 & 17.8 & 11.1 & 49.7 & 73.3 & 39.4 & 35.0 \\
			\rowcolor{red!10} \cellcolor{white} (a) & $50\%$ & $50\%$ & 42.6 & 32.4 & 18.3 & 18.4 & 11.6 & 51.1 & 73.6 & 39.5 & 35.9 \\
			& $75\%$ & $25\%$ & 40.7 & 31.1 & 17.5 & 17.3 & 11.1 & 47.6 & 72.2 & 39.8 & 34.7 \\
			\midrule
			& $25\%$ & $50\%$ & 42.5 & 32.4 & 18.3 & 18.3 & 11.1 & 48.0 & 73.9 & 39.4 & 35.5 \\
			\rowcolor{red!10} \cellcolor{white} (b) & $50\%$ & $50\%$ & 42.6 & 32.4 & 18.3 & 18.4 & 11.6 & 51.1 & 73.6 & 39.5 & 35.9 \\
			& $75\%$ & $50\%$ & 42.8 & 32.6 & 18.4 & 18.4 & 11.9 & 48.3 & 72.7 & 40.1 & 35.6 \\
			\midrule
			& $50\%$ & $25\%$ & 42.4 & 32.3 & 18.3 & 18.3 & 11.5 & 49.9 & 74.1 & 36.5 & 35.4 \\
			\rowcolor{red!10} \cellcolor{white} (c) & $50\%$ & $50\%$ & 42.6 & 32.4 & 18.3 & 18.4 & 11.6 & 51.1 & 73.6 & 39.5 & 35.9 \\
			& $50\%$ & $75\%$ & 43.2 & 33.3 & 18.6 & 18.8 & 11.7 & 51.2 & 73.1 & 39.0 & 36.1 \\
			\bottomrule
		\end{tabular}
	}
	\caption{Comparison of different masking ratios. mIR/mTR refers to the corresponding mean value of R@1, R@5 and R@10.}
	\label{tab:mask ratio}
\end{table}

\subsection{Ablation Study}

We conduct a series of experiments to evalute the impact of training objectives and hyper-parameter settings of SyCoCa. Due to limited computing resources, we train the models on Conceptual Captions 3M (CC3M)~\cite{sharma2018conceptual}, which is a small dataset consisting of filtered image-text pairs. This dataset has been widely used for evaluations of vision-language pretraining~\cite{yang2022unified, zhong2022regionclip, wang2023vilta, dong2023maskclip}.

\textbf{objective Analysis.} To analyze the impact of individual training objectives in SyCoCa, we conduct a series of experiments. We compare the performance of different variants on image-text retrieval and classification tasks, as presented in Table~\ref{tab:object ablation}. We can observe that: 
\begin{itemize}
	\item By incorporating bidirectional cross-modal interaction tasks namely IC and TG-MIM, our proposed SyCoCa achieves an average improvement of $12.5\%$ compared to CoCa on CC3M dataset.
	\item When applying a random masking mechanism, the performance of TG-MIM slightly decreases compared to MIM. That's because the randomly masked patches may not be relevant to the text description, thereby introducing additional noise in the modal interaction and consequently affecting the performance of TG-MIM.
	\item Introducing TG-MIM alone, without the presence of AM, has little impact on performance. This is because attention masking is essential for TG-MIM to rely on text information in order to reconstruct highly correlated visual areas, thereby enhancing the cross-modal interaction between the vision and text modalities. 
	
\end{itemize}

\textbf{Impact of Masking Ratio.} We train SyCoCa with different high/low masking ratios to evaluate their impact on model performance. Our evaluation is designed from three aspects: (a) dividing all image patches into high or low attentive masking, (b) maintaining a fixed low masking ratio while varying the high masking ratio, and (c) vice versa. The results are presented in Table~\ref{tab:mask ratio}. From Table~\ref{tab:mask ratio}~(a), we can observe that dividing the image patches evenly yields better results compared to biased partitioning, as it balances the visual information used in objectives in terms of image captioning and Tg-MIM. Moreover, in Table~\ref{tab:mask ratio}~(b)-(c) SyCoCa is not sensitive to changes in the mask ratio. 

\textbf{Training objective Weights.} We also investigate the impact of different weights for the TG-MIM, by varying the weights $\lambda_{TM}$ assigned to the TG-MIM loss while keeping other objectives' weights fixed. The results in Table~\ref{tab:weight ablation} indicate that within the specific range of $[0.1, 1.0]$, the variation of $\lambda_{TM}$ has minimal impact on the performance. We select $\lambda_{TM}=1.0$ for training our model.

\begin{table}[]
	\centering
	\resizebox{\linewidth}{!}{%
		\begin{tabular}{l|cccc|cccc|c}
			\toprule
			& \multicolumn{4}{c|}{zero-shot retrieval} & \multicolumn{4}{c|}{zero-shot classification} \\
			\cmidrule(r){2-5} \cmidrule(r){6-9}
			& \multicolumn{2}{c}{Flickr30K} & \multicolumn{2}{c|}{MSCOCO} & \multirow{2}{*}{IN-1K} & \multirow{2}{*}{C-10} & \multirow{2}{*}{STL-10} & \multirow{2}{*}{Caltech} & \\
			\cmidrule(r){2-3} \cmidrule(r){4-5}
			$\lambda_{TM}$ & mTR & mIR & mTR & mIR &  &  &  & & Avg. \\
			\midrule
			$0.1$ & 42.1 & \textbf{32.4} & \textbf{18.3} & 18.2 & 11.3 & 46.5 & 73.2 & 39.8 & 35.2 \\
			$0.5$ & \textbf{42.8} & 32.2 & 18.2 & 18.2 & 11.5 & 46.1 & \textbf{73.6} & 38.6 & 35.1 \\
			\rowcolor{red!10} $1.0$ & 42.6 & \textbf{32.4} & \textbf{18.3} & \textbf{18.4} & \textbf{11.6} & \textbf{51.1} & \textbf{73.6} & 39.5 & \textbf{35.9} \\
			$2.0$ & 40.5 & 31.3 & 17.7 & 17.8 & 11.3 & 47.3 & 72.2 & \textbf{40.0} & 34.8 \\
			\bottomrule
		\end{tabular}%
	}
	\caption{Comparison of different weights for the TG-MIM objective. mIR/mTR refers to the corresponding mean value of R@1, R@5 and R@10.}
	\label{tab:weight ablation}
\end{table}

\section{Conclusion}

In this paper, we introduce a novel vision-language pretraining method called SyCoCa, which aims to further enhance multi-modal alignment. Our approach focuses on improving the fine-grained understanding between vision and language modalities by introducing the text-guided masked image modeling (TG-MIM) training objective. By incorporating the TG-MIM training objective into the CoCa framework, we establish bidirectional local interaction, which leads to a precise and fine-grained alignment between the vision and language modalities. Additionally, we propose a new attentive masking approach for TG-MIM, which selectively masks image patches that have strong correlations with the text caption. By focusing on these highly relevant patches, we enhance the cross-modal interaction and significantly improve overall performance. Through extensive experiments on five vision-language tasks, we demonstrate the effectiveness and generalization ability of our SyCoCa.

{
	\small
	\bibliographystyle{ieeenat_fullname}
	\bibliography{arxiv}
}

\newpage

\appendix
\section*{Appendix: Comparison with other methods}
To verify the effectiveness of our proposed Text-Guided Masked Image Modeling (TG-MIM) and attentive masking (AM) strategy, the experiments are conducted under the same settings of CoCa and SyCoCa in the main paper. Especially for the training dataset, we used CC12M, a small dataset with 12 million image-text pairs. As we all know, the amount of training dataset has a crucial impact on model performance. To eliminate data factors and reflect the superiority of our algorithm, we use a large-scale image-text pairs dataset for pre-training. Specifically, we collect Laion-2B \cite{schuhmann2022laion} and COYO-700M \cite{kakaobrain2022coyo-700m}. In addition, the number of parameters will also affect the model performance. To train the collected datasets, we use a total parameter size of 930M. Tab \ref{tab:model configurations sup} summarizes the pre-training 
 model configurations of the related methods. The comparison results between our SyCoCa and other methods are shown in Tab \ref{tab:zero-shot retrieval sup} and Tab \ref{tab:zero-shot caption sub}.

\textbf{Image-Text Retrieval}. For image-text retrieval task, the comparison results are shown in Tab \ref{tab:zero-shot retrieval sup}. Our SyCoCa ahieves the best results in terms of text-to-image retrieval on the Flickr30K dataset, as well as text-to-image/image-to-text retrieval on the MSCOCO dataset. It is worth noting that, SyCoCa outperforms ALIGN~\cite{jia2021scaling} in 11 out of 12 cases, despite both models having similar model parameter sizes and pre-training dataset sizes.
In addition, even though some models (such as CoCa, BEiT-3, and SigLIP) employ larger parameter sizes and datasets, our SyCoCa still achieves comparable results. For instance, the mean text-to-image/image-to-text recall results for SyCoCa are $96.0/90.5/\textbf{82.3}/\textbf{69.8}$, while for CoCa, they are $\textbf{97.3}/\textbf{91.3}/81.4/69.1$.

\textbf{Image Caption}. For image caption task, the comparison results are shown in  Tab \ref{tab:zero-shot caption sub}. On MSCOCO dataset, our SyCoCa is slightly lower than the best method on BLEU@4 and SPICE, but outperforms CoCa on all metrics. On NoCaps dataset, our method achieves the best performance, which shows that our attentive mask strategy has certain advantages in the caption task.

\begin{table*}[t]
\centering
\begin{tabular}{lcccc}
\toprule
Model & total \#param. & precision & dataset & image size \\
\midrule
CLIP~\cite{radford2021learning}         & 430M  & \texttt{fp16}  & WIT-400M     & $336^2$ \\
OpenCLIP~\cite{cherti2022reproducible}  & 430M  & \texttt{bf16}  & LAION-2B     & $224^2$ \\
ALIGN~\cite{jia2021scaling}             & 820M  & -              & ALIGN-1.8B   & $289^2$ \\
FILIP~\cite{yao2021filip}               & 430M  & \texttt{fp16}  & FILIP-340M   & $224^2$ \\
Florence~\cite{yuan2021florence}        & 890M  & -              & FLD-900M     & $224^2$ \\
CoCa~\cite{yu2022coca}                  & 2.1B  & -              & JFT-3B+ALIGN-1.8B       & $288^2$ \\
CoCa-Large~\cite{yu2022coca}            & 790M  & -              & JFT-3B+ALIGN-1.8B       & $288^2$ \\
BEiT-3~\cite{wang2022image}             & 1.9B  & -              & 15M images+160GB documents+21M pairs & $224^2$ \\
EVA-CLIP~\cite{sun2023eva}              & 430M  & -              & LAION-2B+COYO-700M & $224^2$ \\
SigLIP~\cite{zhai2023sigmoid}           & 430M  & -              & WebLI$^\dagger$              & $224^2$ \\ 
CLIPA-v2~\cite{li2023clipav2}           & 1.0B  & -              & DataComp-1B        & $336^2$ \\
LiT~\cite{zhai2022lit}                  & 1.3B  & -              & LiT-4B       & $288^2$ \\
\midrule
SyCoCa (ours)                           & 930M  & \texttt{bf16}  & LAION-2B+COYO-700M & $224^2$ \\
\bottomrule
\end{tabular}
\caption{Pre-training model configurations. $^\dagger$ WebLI is a private multilingual dataset conducted by Google, which consists of 10 billion images and 12 billion alt-texts.}
\label{tab:model configurations sup}
\end{table*}

\begin{table*}[t]
\centering
\begin{tabular}{lcccccccccccc}
\toprule
& \multicolumn{6}{c}{Flickr30K} & \multicolumn{6}{c}{MSCOCO} \\
\cmidrule(r){2-7} \cmidrule(r){8-13}
& \multicolumn{3}{c}{Image $\rightarrow$ Text} & \multicolumn{3}{c}{Text $\rightarrow$ Image} & \multicolumn{3}{c}{Image $\rightarrow$ Text} & \multicolumn{3}{c}{Text $\rightarrow$ Image} \\
\cmidrule(r){2-4} \cmidrule(r){5-7} \cmidrule(r){8-10} \cmidrule(r){11-13}
Model & R@1 & R@5 & R@10 & R@1 & R@5 & R@10 & R@1 & R@5 & R@10 & R@1 & R@5 & R@10 \\
\midrule
CLIP~\cite{radford2021learning}              & 88.0 & 98.7 & 99.4 & 68.7 & 90.6 & 95.2 & 58.4 & 81.5 & 88.1 & 37.8 & 62.4 & 72.2 \\
OpenCLIP~\cite{cherti2022reproducible}       & 88.7 & 98.4 & 99.2 & 75.0 & 92.5 & 95.6 & 62.1 & 83.4 & 90.3 & 46.1 & 70.7 & 79.4 \\
ALIGN~\cite{jia2021scaling}                  & 88.6 & 98.7 & \underline{99.7} & 75.7 & \underline{93.8} & \underline{96.8} & 58.6 & 83.0 & 89.7 & 45.6 & 69.8 & 78.6 \\
FILIP~\cite{yao2021filip}                    & \underline{89.8} & \textbf{99.2} & \textbf{99.8} & 75.0 & 93.4 & 96.3 & 61.3 & 84.3 & \underline{90.4} & 45.9 & 70.6 & 79.3 \\
Florence~\cite{yuan2021florence}             & \textbf{90.9} & \underline{99.1} & -    & 76.7 & 93.6 & -    & 64.7 & \underline{85.9} & -    & 47.2 & \underline{71.4} & -    \\
\textcolor{gray}{CoCa~\cite{yu2022coca}}      & \textcolor{gray}{92.5} & \textcolor{gray}{99.5} & \textcolor{gray}{99.9} & \textcolor{gray}{80.4} & \textcolor{gray}{95.7} & \textcolor{gray}{97.7} & \textcolor{gray}{66.3} & \textcolor{gray}{86.2} & \textcolor{gray}{91.8} & \textcolor{gray}{51.2} & \textcolor{gray}{74.2} & \textcolor{gray}{82.0} \\
\textcolor{gray}{BEiT-3~\cite{wang2022image}} & \textcolor{gray}{94.9} & \textcolor{gray}{99.9} & \textcolor{gray}{100.0}& \textcolor{gray}{81.5} & \textcolor{gray}{95.6} & \textcolor{gray}{97.8} & \textcolor{gray}{-} & \textcolor{gray}{-} & \textcolor{gray}{-} & \textcolor{gray}{-} & \textcolor{gray}{-} & \textcolor{gray}{-} \\ 
EVA-CLIP~\cite{sun2023eva}             & 89.7 & 98.6 & 99.2 & \underline{77.3} & 93.6 & \underline{96.8} & 63.7 & 84.3 & \underline{90.4} & \underline{47.5} & 71.2 & \underline{79.7} \\
\textcolor{gray}{SigLIP~\cite{zhai2023sigmoid}}& \textcolor{gray}{-}    & \textcolor{gray}{-}    & \textcolor{gray}{-}    & \textcolor{gray}{-}    & \textcolor{gray}{-}    & \textcolor{gray}{-}    & \textcolor{gray}{70.2} & \textcolor{gray}{-}    & \textcolor{gray}{-}    & \textcolor{gray}{52.0} & \textcolor{gray}{-}    & \textcolor{gray}{-}    \\
CLIPA-v2~\cite{li2023clipav2}                & 89.1 & -    & -    & 73.0 & -    & -    & 64.1 & -    & -    & 46.3 & -    & -    \\
\midrule
SyCoCa (ours)                                & 89.2 & \underline{99.1} & 99.6 & \textbf{78.7} & \textbf{95.4} & \textbf{97.4} & \textbf{67.2} & \textbf{87.5} & \textbf{92.1} & \textbf{50.7} & \textbf{75.7} & \textbf{82.9} \\
\bottomrule
\end{tabular}
\caption{Zero-shot image-text retrieval comparisons on Flickr30K~\cite{flickr30k} and MSCOCO~\cite{mscoco}. Results of models that use significantly larger parameter sizes or dataset sizes are indicated in \textcolor{gray}{gray}.}
\label{tab:zero-shot retrieval sup}
\end{table*}

\begin{table*}[t]
\centering
\begin{tabular}{lcccccccc}
\toprule
& \multicolumn{4}{c}{MSCOCO} & \multicolumn{4}{c}{NoCaps} \\
\cmidrule(r){2-5} \cmidrule(r){6-9}
& & & & & \multicolumn{2}{c}{val} & \multicolumn{2}{c}{test} \\
\cmidrule(r){6-7} \cmidrule(r){8-9}
Model & B@4 & M & C & S & C & S & C & S \\
\midrule
CLIP-ViL~\cite{shen2021much}                & 40.2 & 29.7 & 134.2 & 23.8 & -     & -    & -     & -    \\
BLIP~\cite{li2022blip}                      & 40.4 & -    & 136.7 & -    & 113.2 & 14.8 & -     & -    \\
VinVL~\cite{zhang2021vinvl}                 & 41.0 & 31.1 & 140.9 & \textbf{25.4} & 105.1 & 14.4 & 103.7 & 14.4 \\
SimVLM~\cite{wang2021simvlm}                & 40.6 & 33.7 & 143.3 & \textbf{25.4} & 112.2 & -    & 110.3 & 14.5 \\
LEMON~\cite{hu2022scaling}                  & \textbf{41.5} & 30.8 & 139.1 & 24.1 & 117.3 & 15.0 & 114.3 & 14.9 \\
CoCa~\cite{yu2022coca}                      & 40.9 & 33.9 & \underline{143.6} & 24.7 & \underline{122.4} & \underline{15.5} & \underline{120.6} & \underline{15.5} \\
\midrule
SyCoCa (ours) & \underline{41.4} & \textbf{34.1} & \textbf{143.7} & \underline{25.3} & \textbf{122.6} &\textbf{15.8} & \textbf{121.1} & \textbf{15.6} \\ 

\bottomrule
\end{tabular}
\caption{Image Captioning comparisons on MSCOCO~\cite{mscoco} and NoCaps~\cite{nocaps}. B@4: BLEU@4, M: METEOR, C: CIDEr, S: SPICE.}
\label{tab:zero-shot caption sub}
\end{table*}


\end{document}